\documentclass[9pt, letterpaper]{ifacconf}
\usepackage[left=1.5cm, right=1.5cm, top=3.3cm, bottom=1.5cm]{geometry}
\usepackage{mathptmx}
\usepackage{graphicx}      
\usepackage[sort&compress,numbers]{natbib}        
\usepackage[font=footnotesize, center]{caption}

\usepackage{import}
\usepackage{multirow}
\usepackage{units}

\usepackage{fancyhdr}
\usepackage{tikz, pgfplots}
\usetikzlibrary{arrows, positioning, patterns, shapes, shadows, decorations.pathreplacing, backgrounds, fit, hobby, decorations.markings}
				
\usepackage{amsmath}
\usepackage{pgfplotstable}
\usepgfplotslibrary{groupplots}

\usepackage[pscoord]{eso-pic}%
\newcommand{\placetextbox}[3]{
  \setbox0=\hbox{#3}
  \AddToShipoutPictureFG*{
    \put(\LenToUnit{#1\paperwidth},\LenToUnit{#2\paperheight}){\vtop{{\null}\makebox[0pt][c]{#3}}}%
  }%
}%


\usepackage{changepage}

\usepackage{titlesec}

\titleformat*{\section}{\bfseries \centering}
\titleformat*{\subsection}{\normalfont }
\titleformat*{\subsubsection}{\normalfont }
\titleformat*{\paragraph}{\normalfont }
\titleformat*{\subparagraph}{\normalfont }

\makeatletter
\def\tagform@#1{\maketag@@@{(#1)\unskip\@@italiccorr}\hspace{1em}}
\makeatother

\makeatletter
\renewcommand{\@seccntformat}[1]{\csname the#1\endcsname.\quad}
\makeatother

\usepackage{epstopdf}

\newlength\figureheight
\newlength\figurewidth

\pgfplotsset{ every axis label/.append style={font=\scriptsize},
every axis y label/.append style={yshift=-0.5cm},
every axis x label/.append style={yshift=0.2cm},
every tick label/.append style={font=\scriptsize}, compat=newest}

\usepackage{comment}

\begin{document}

\newcommand{\doublespace}{}

\newcommand{\afz}[1]{``#1''}
\newcommand{\afzsimple}[1]{`#1'}

\newcommand{\beq}{\begin{equation}}
\newcommand{\eeq}{\end{equation}}
\newcommand{\beqo}{\begin{displaymath}}
\newcommand{\eeqo}{\end{displaymath}}
\newcommand{\bea}{\begin{eqnarray}}
\newcommand{\eea}{\end{eqnarray}} 
\newcommand{\beao}{\begin{eqnarray*}}
\newcommand{\eeao}{\end{eqnarray*}}

\newcommand{\er}[1]{eq.(\ref{#1})}
\newcommand{\erwo}[1]{(\ref{#1})}

\newcommand{\norm}[1]{\left|\!\left| #1 \right|\!\right|}
\newcommand{\bignorm}[1]{\big|\!\big| #1 \big|\!\big|}
\newcommand{\mean}[1]{\left\langle #1 \right\rangle}
\newcommand{\doublemean}[1]{\left\langle\left\langle #1 \right\rangle\right\rangle}
\newcommand{\braquet}[2]{\left\langle #1 \bigm| #2 \right\rangle}

\newcommand{\Nfat}{{\rm I\hspace{-0.23em}N}}
\newcommand{\Rfat}{{\rm I\hspace{-0.23em}R}}
\newcommand{\Zfat}{{\rm Z}}

\newcommand{\ddt}[1]{\frac{\mbox{d}#1}{\mbox{d}t}}
\newcommand{\smallddt}[1]{{\mbox{d}#1}/{\mbox{d}t}}
\newcommand{\dds}[1]{\frac{\mbox{d}#1}{\mbox{d}s}}
\newcommand{\intd}[1]{\mbox{d}#1}
\newcommand{\smalldds}[1]{{\mbox{d}#1}/{\mbox{d}s}}
\newcommand{\Deltat}{{\Delta t}}
\newcommand{\deltat}{{\scriptsize \Delta}t}
\newcommand{\Deltaxvec}{{\Delta \xvec}}
\newcommand{\deltaxvec}[1]{{\scriptsize \Delta}\xvec}


\newcommand{\avec}{{\mathbf{a}}}
\newcommand{\cvec}{{\mathbf{c}}}
\newcommand{\dvec}{{\mathbf{d}}}
\newcommand{\evec}{{\mathbf{e}}}
\newcommand{\fvec}{{\mathbf{f}}}
\newcommand{\gvec}{{\mathbf{g}}}
\newcommand{\kvec}{{\mathbf{k}}}
\newcommand{\lvec}{{\mathbf{l}}}
\newcommand{\mvec}{{\mathbf{m}}}
\newcommand{\nvec}{{\mathbf{n}}}
\newcommand{\pvec}{{\mathbf{p}}}
\newcommand{\rvec}{{\mathbf{r}}}
\newcommand{\tvec}{{\mathbf{t}}}
\newcommand{\vvec}{{\mathbf{v}}}
\newcommand{\wvec}{{\mathbf{w}}}
\newcommand{\xvec}{{\mathbf{x}}}
\newcommand{\yvec}{{\mathbf{y}}}
\newcommand{\zvec}{{\mathbf{z}}}

\newcommand{\thetavec}{{\mathbf{\theta}}}

\newcommand{\Avec}{{\mathbf{A}}}
\newcommand{\Bvec}{{\mathbf{B}}}
\newcommand{\Cvec}{{\mathbf{C}}}
\newcommand{\Dvec}{{\mathbf{D}}}
\newcommand{\Evec}{{\mathbf{E}}}
\newcommand{\Fvec}{{\mathbf{F}}}
\newcommand{\Gvec}{{\mathbf{G}}}
\newcommand{\Hvec}{{\mathbf{H}}}
\newcommand{\Ivec}{{\mathbf{I}}}
\newcommand{\Jvec}{{\mathbf{J}}}
\newcommand{\Kvec}{{\mathbf{K}}}
\newcommand{\Lvec}{{\mathbf{L}}}
\newcommand{\Mvec}{{\mathbf{M}}}
\newcommand{\Nvec}{{\mathbf{N}}}
\newcommand{\Ovec}{{\mathbf{O}}}
\newcommand{\Pvec}{{\mathbf{P}}}
\newcommand{\Qvec}{{\mathbf{Q}}}
\newcommand{\Rvec}{{\mathbf{R}}}
\newcommand{\Svec}{{\mathbf{S}}}
\newcommand{\Tvec}{{\mathbf{T}}}
\newcommand{\Uvec}{{\mathbf{U}}}
\newcommand{\Vvec}{{\mathbf{V}}}
\newcommand{\Wvec}{{\mathbf{W}}}
\newcommand{\Xvec}{{\mathbf{X}}}
\newcommand{\Yvec}{{\mathbf{Y}}}
\newcommand{\Zvec}{{\mathbf{Z}}}
\newcommand{\Zerovec}{{\mathbf{0}}}
\newcommand{\Onevec}{{\mathbf{1}}}

%

\newcommand{\spodot}{{\!\odot}}
\newcommand{\alphaCompExp}{\alpha \hspace{-0.22cm} \bigcirc}
\newcommand{\twoCompExp}{\mbox{\tiny{$2$}} \hspace{-0.19cm} \bigcirc}

\newcommand{\vext}{v^{\mbox{\tiny ext}}}
\newcommand{\vsyn}{h}
\newcommand{\vrest}{v^{\mbox{\tiny Rest}}}
\newcommand{\vref}{v^{\mbox{\tiny ref}}}
\newcommand{\etasingle}{\eta^{\mbox{\tiny s}}}
\newcommand{\ethreemin}{\eta_3^{\mbox{\tiny min}}}
\newcommand{\etwomin}{\eta_2^{\mbox{\tiny min}}}

\newcommand{\nospikes}{N^{\mbox{\tiny burst}}}
\newcommand{\noburstspikes}{n^{\mbox{\tiny burst}}}


\newcommand{\tauref}{\tau_{\mbox{\tiny ref}}}
\newcommand{\taualpha}{\tau_\alpha}
\newcommand{\tausrm}{\tau^{\mbox{\tiny SRM}}}
\newcommand{\tauesc}{\tau^{\mbox{\tiny esc}}}
\newcommand{\taugraded}{\tau_{\mbox{\tiny g}}}
\newcommand{\gabs}{\gamma^{\mbox{\tiny abs}}}
\newcommand{\Dax}{\Delta^{\mbox{\tiny ax}}}

\newcommand{\Amax}{A_{\mbox{\tiny max}}}
\newcommand{\smax}{s_{\mbox{\tiny max}}}

\newcommand{\sigmastat}{\sigma_{\mbox{\tiny stat}}}
\newcommand{\rstat}{r_{\mbox{\tiny stat}}}


\newcommand{\pa}{p_{\mbox{\tiny A}}}
\newcommand{\Pa}{P_{\mbox{\tiny A}}}
\newcommand{\NI}{N_{\mbox{\tiny I}}}
\newcommand{\Gammaa}{\Gamma_{\mbox{\tiny A}}}

\newcommand{\jd}{j^{\mbox{\tiny drift}}}
\newcommand{\jcrit}{J^{\mbox{\tiny crit}}}
\newcommand{\jeff}{J^{\mbox{\tiny eff}}}
\newcommand{\jext}{J^{\mbox{\tiny ext}}}
\newcommand{\jself}{J^{\mbox{\tiny self}}}
\newcommand{\jsingle}{J^{\mbox{\tiny single}}}
\newcommand{\jcoup}{J^{\mbox{\tiny coup}}}
\newcommand{\jfeed}{J^{\mbox{\tiny feed}}}

\newcommand{\etastandard}{\eta^1}
\newcommand{\Nspikes}{N_{\mbox{\tiny spikes}}}

\newcommand{\sumjm}{\sum_{j=1}^{l_m}}
\newcommand{\sumjmlock}{\sum_{j=1,j\; {\mbox{\scriptsize locks}}}^{l_m}}
\newcommand{\sumjmdrift}{\sum_{j=1,j\; {\mbox{\scriptsize drifts}}}^{l_m}}
\newcommand{\sumjmspikes}{\sum_{j=1,j\; {\mbox{\scriptsize spikes}}}^{l_m}}
\newcommand{\sumpoolsn}{\sum_{n=1}^{L}}
\newcommand{\sumpoolsnwom}{\sum_{n=1,n\ne m}^{L}}

\newcommand{\rhol}{\rho^{\mbox{\tiny lock}}}
\newcommand{\rhod}{\rho^{\mbox{\tiny drift}}}
\newcommand{\rhof}{\rho^{\mbox{\tiny field}}}

\newcommand{\Amean}{A^{\mbox{\tiny mean}}}
\newcommand{\Alock}{A^{\mbox{\tiny lock}}}
\newcommand{\Adrift}{A^{\mbox{\tiny drift}}}

\newcommand{\plock}{p^{\mbox{\tiny lock}}}
\newcommand{\pdrift}{p^{\mbox{\tiny drift}}}

\newcommand{\rtotal}{r^{\mbox{\tiny total}}}
\newcommand{\Psitotal}{\Psi^{\mbox{\tiny total}}}
\newcommand{\Atotal}{A^{\mbox{\tiny total}}}
\newcommand{\Stotal}{S^{\mbox{\tiny total}}}
\newcommand{\jsimple}{J^{\mbox{\tiny feed,simple}}}
\newcommand{\jintra}{J^{\mbox{\tiny feed,intra}}}
\newcommand{\jinter}{J^{\mbox{\tiny feed,inter}}}
\newcommand{\jinterincl}{J^{\mbox{\tiny feed,inter,incl}}}
\newcommand{\jinterexcl}{J^{\mbox{\tiny feed,inter,excl}}}
\newcommand{\jSD}{J^{\mbox{\tiny SD}}}
\newcommand{\jCD}{J^{\mbox{\tiny CD}}}

\newcommand{\tauup}{\tau^{\mbox{\tiny up}}}
\newcommand{\taudown}{\tau^{\mbox{\tiny down}}}
\newcommand{\rsp}{r^{\mbox{\tiny SP}}}
\newcommand{\rog}{r^{\mbox{\tiny OG}}}
\newcommand{\Ssp}{S^{\mbox{\tiny SP}}}
\newcommand{\Sog}{S^{\mbox{\tiny OG}}}

\begin{frontmatter}

\raisebox{2.75cm}[3cm][0cm]{\hspace{-10cm}\includegraphics[width=0.2\textwidth]{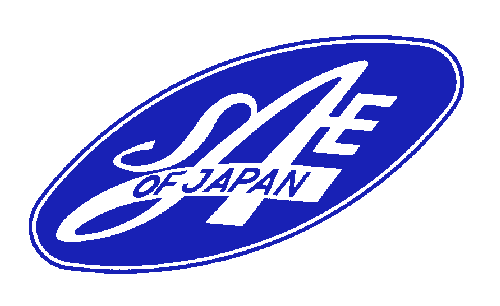}}
\newline
\raisebox{5.7cm}[3cm][0cm]{\hspace{10.3cm}}
\vspace{-6.8cm}

\title{\mbox{\normalfont Continuous Risk Measures for Driving Support}}

\vspace{0.4cm}

\hspace*{-2.1cm} \begin{minipage}[15cm]{15cm}
\textbf{\centerline {Julian Eggert\textsuperscript{1)} Tim Puphal\textsuperscript{1)}}}

\vspace{0.27cm}
\textit{\centerline{1) Honda Research Institute (HRI) Europe}
\textit{\centerline{Carl-Legien-Str. 30, 63073 Offenbach, Germany (e-mail:[julian.eggert, tim.puphal]@honda-ri.de)}}
}

\vspace{0.11cm}
\centerline{Received on November 17, 2017} 

\vspace{0.6cm}

\textbf{ABSTRACT}: In this paper, we compare three different model-based risk measures by evaluating their stengths and weaknesses qualitatively and 
testing them quantitatively on a set of real longitudinal and intersection scenarios. 
We start with the traditional heuristic Time-To-Collision (TTC), which we extend 
towards 2D operation and non-crash cases to retrieve the Time-To-Closest-Encounter (TTCE). The second risk measure models position uncertainty with a Gaussian distribution
and uses spatial occupancy probabilities for collision risks. We then derive a novel risk measure based on the statistics of sparse critical events and 
so-called \afz{survival} conditions. The resulting survival analysis shows to have an earlier detection time
of crashes and less false positive detections in near-crash and non-crash cases supported by its solid theoretical grounding. 
It can be seen as a generalization of TTCE and the Gaussian method which is suitable for the validation of ADAS and AD. 
\vspace{0.06cm}

\end{minipage}

\end{frontmatter}


\section{Introduction}\label{SecIntroduction}

\placetextbox{0.5}{0.018}{Copyright \copyright \ 2018 Society of Automotive Engineers of Japan, Inc. All rights reserved}

\vspace{-2mm}

Current progress in the development of Advanced Driver Assistance Systems (ADAS) and Autonomous Driving (AD) is 
based on a broad range of technological and methodological advances in the field of artificial intelligence. 
One objective of future ADAS and AD is the improvement of road safety by foreseeing dangerous situations and supporting the driver to behave in an appropriate 
way mitigating accidents. A key component is therefore the capability of situation risk assessment.
Risk is usually defined as the likelihood that a critical event might occur weighted by its probable severity, i.e. its 
potential consequences in terms of damage, costs and injuries\textsuperscript{ (1)}. 
The main problem with risk estimation is its predictive character involving uncertainties 
(in sensor measurements, driver behavior and scene evolution), which spread over time and which have to be modeled in 
a sensible way.

\hspace{0.4cm} In previous related work, numerous approaches for prediction and risk assessment have been introduced\textsuperscript{ (2)}.
On the one hand, it is possible to calculate risks by detecting hazardous driver intentions. 
For example, the typical steering behavior is learnable using image streams in a 
with convolutional neural networks\textsuperscript{ (3)}. Comparing the measured with the learned wheel angle, the deviation can be assumed to correlate to the current risk. Since only the road structure influences the nominal behavior, the resulting risk value indicates curve risk and not collision risk. 
As an alternative, a Bayesian network\textsuperscript{ (4)} is employed to classify behaviors into typical maneuvers, e.g. drive straight or turn. Training the network with accident data would lead to maneuver detections of vehicles violating traffic rules. Hereby, risks cannot be identified for situations which are not in the dataset.

\hspace{0.4cm} On the other hand, risk measures are based on future vehicle trajectories. The Time-To-Collision (TTC)\textsuperscript{ (5)}, as one example of Time-To-Event (TTX) indicators, is defined as the deterministic time until the trajectories of two vehicles  intersect. For the trajectory prediction, a constant velocity model is implicitly given in the respective equations. Variants of TTC incorporate different velocity profiles, such as constant acceleration models\textsuperscript{ (6)}. Since TTC only works for longitudinal scenarios, its equations have been extended for 2D operation\textsuperscript{ (7)}. 
Additionally, to overcome the collision assumption in TTC 
and thus having realistic values in non-crash scenarios the orientation of the vehicles are taken into account. 
Both drawbacks of TTC are also handled directly with the Time-To-Closest-Encounter (TTCE)\textsuperscript{ (8)}.

\begin{figure*}[t!]
	\centering
 	\resizebox{0.7\linewidth}{!}{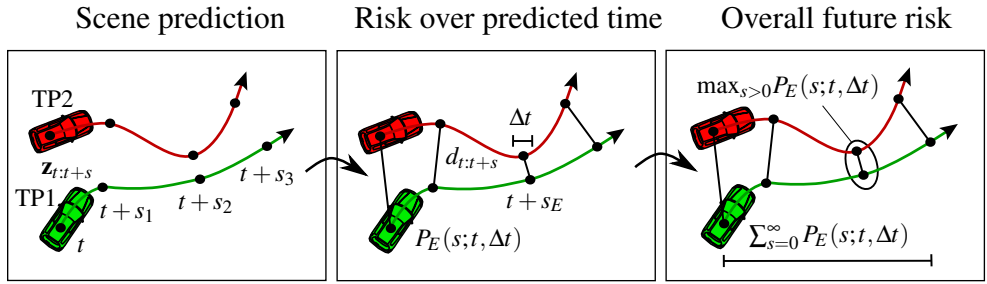}
 	\captionsetup{justification=centering}
	\caption{\centering General approach for collision risk prediction of two traffic participants.}
	\label{fig:framework}
     \vspace{0.1cm}
\end{figure*}

\hspace{0.4cm} Probabilistic risk prediction presumes critical events to act in a defined probability distribution. 
Gaussian methods\textsuperscript{ (9)} model a normal distribution for the positions in the predicted 
trajectories and calculate the collision probability by estimating their overlap. 
Furthermore to improve the accuracy, the distribution variances are separated into longitudinal and lateral
components and instead of integrating analytically, Monte Carlo simulation gives an approximation of the overlap\textsuperscript{ (10)}. 
Similarly, the concept of Kamms circle\textsuperscript{ (11)} is used as positional uncertainty to derive a ``Worst-TTC'' as the maximal risk value. With Poisson processes\textsuperscript{ (12)} the probability to remain accident-free can be calculated out of the mean time between critical events, which is dependent on the distances of the future trajectories. 
The risk measure follows as the complement of the corresponding survival function 
and allows to consider not only collision risk, but also for instance the risk for losing control in curves.

\hspace{0.4cm} Despite these works, a theoretically grounded, generalizable, yet efficient risk measure for ADAS and AD is still missing.
Risks on the behavioral level\textsuperscript{ (3,4)} are based on machine learning algorithms and their accuracy is
highly influenced by the available data. In this paper, we therefore derive the model-based risk measures 
TTCE, Gaussian method and survival analysis and compare them in simulations with the purpose of analyzing their 
properties for an application to ADAS and AD. Each measure is a representative of a broader family\textsuperscript{ (8,9,12)} and has been enhanced
so that they can deal with temporal and spatial uncertainties. The depicted results are based on a conference paper
published in FAST-zero 2017.\textsuperscript{ (13)}
In the next Section \ref{SecRequirements}, we outline the properties of a suitable risk measure and
the general steps for its computation. Section \ref{SecTheoRisk} gives a detailed introduction of the three 
chosen risk measures and in Section \ref{SecSimRes} we show their performance and robustness on a set of
longitudinal and intersection scenarios. 
Finally, in Section \ref{SecDiscOut} we discuss future possible research areas.

\section{Requirements and Framework} \label{SecRequirements}

\hspace{0.4cm} As a starting point, we consider dynamic collision scenarios (i.e. a traffic scene with two traffic participants $\mbox{TP}1, \mbox{TP}2$) at an 
arbitrary moment in time $t$. Beginning at $t$, the target is to estimate the risk of a critical event that could happen at a 
future time $t+s$, that is at a temporal distance $s$ into the future. We assume the events to be disruptive and to have no 
duration, so that they can be fully characterized by their time $t+s$, if they happen at $s$ into the future.

\hspace{0.4cm} Since most of the commonly used risk measures do not address severity explicitly,
we will concentrate on risk as an event occurrence probability. However, severity can
be included into the argumentation in a straightforward way.
An indicator for risk is then the probability function $P_E(s;t,\Delta t)$ that a critical event will happen during an interval 
of size $\Delta t$ around a future time $t+s$. 
As a probability it should be positive and appropriately normalized, so that $P_E(s;t,\Delta t)\in [0,1]$. With a collision of time $t_E$ (time {\em until} 
a collision occurs $s_E$), if a collision is imminent (at $s_E\rightarrow 0$), it should be $\lim_{s \rightarrow 0} \,P_E(s;t,\Delta t) 
\rightarrow 1$ and if no collision ever occurs, it should be $\lim_{s \rightarrow \infty} \,P_E(s;t,\Delta t) \rightarrow 0$.

\hspace{0.4cm} A compact risk measure $R(t)$ would comprise, for each point in time $t$, the entire {\bf accumulated expected future risk} 
contained in $P_E(s;t,\Delta t)$, $s\in[0,\infty]$. There are several possible ways to gain such a measure, e.g.~by extracting 
the maximal expected risk
\beq\label{EqRMax}
R(t):=\mbox{max}_{s>0} \, P_E(s;t,\Delta t)
\eeq

or using an appropriatedly accumulated risk
\beq\label{EqRInt}
R(t):=\sum_{s=0}^{\infty} P_E(s;t,\Delta t) \ \ .
\eeq
The latter is a cleaner form because it comprises the full future event probability, but it requires a more careful 
derivation due to proper normalization considerations. Additionally, heuristically motivated risk measures often 
directly estimate $R(t)$ without $P_E(s;t,\Delta t)$ (see Section \ref{SecRMTTCE}). 
Again here, it should be guaranteed that $R(t)\in[0,1]$ and $R(t)\rightarrow 1$ for collision at $s_E\rightarrow 0$ and $R(t)<1$ 
for no collision at any time.

\hspace{0.4cm} A risk estimation framework consists of three components as depicted in Fig. \ref{fig:framework}. In a first step, a prediction of how 
the situation will evolve in the future is performed. In our notation, designating $\zvec$ as the state vector of a 
scene, the predicted sequence of future scene states is given by $\zvec_{t:t+s}$. 
The prediction can thereby be modeled at different levels of detail depending on the geometry, kinematics and interaction. For example, 
a low-level prediction would treat vehicles as dynamic entities and use constant acceleration, 
velocity or turn-rate assumptions in kinematic equations. On a next level, road geometries could be taken into account to constrain the paths on which 
vehicles can drive. The outcome of the prediction is a selected set of possible future trajectories for each involved vehicle.

\hspace{0.4cm} In a second step, the time evolution of the scene $\zvec_{t:t+s}$ is evaluated in terms of criticality by extracting
features which are indicative of risk. 
For collision risk, the predicted trajectories are compared for each point in predicted time $s$ to 
obtain the spatiotemporal proximity $d_{t:t+s}$ between the vehicle trajectories. 
This leads to an {\bf instantaneous risk} function or in features like the time to and distance at the point of maximal risk, 
resp.~the time until 
the event $s_E$ and the predicted proximity $d_{t+s_E}$ at that time. Afterwards, the event probability $P_E(s;t,\Delta t)$ can be 
calculated from the instantaneous risk function. The third step comprises retrieving the risk measure in form of a scalar risk function 
$R(t)$ according to Eq. (1) or (2). \\
\vspace{-0.15cm}
\section{Theory of Risk Measures} \label{SecTheoRisk}

\subsection{Time-To-Closest-Encounter (TTCE)}\label{SecRMTTCE}


\hspace{0.4cm} The family of TTX-based risk measures are proximal safety indicators based on the time left until a critical event. 
In particular, the well-known TTC represents the time remaining
until two vehicles will engage in a collision if they continue driving along the same path according to some prediction model\textsuperscript{ (5)}. 
A usual assumption is that both TP's drive with constant longitudinal velocities $v_{1,2}$. In this case, if they start driving 
at $t$ with longitudinal positions $l_{1,2}$, the time of collision / time of critical event will be at $s_E=\mbox{TTC}=-\Delta l/\Delta v$, 
with $\Delta l:=l_1-l_2$ and $\Delta v = v_1-v_2$. 
However, TTC is rather limited for complex scenes, 
because (i) it is only applicable to longitudinal resp.~1D scenarios and (ii) it presumes that a collision will happen with certainty, 
so that near-crash cases cannot be evaluated. 
Several extensions of TTC to 2D 
have been developed\textsuperscript{ (7)}, but generally lack justification both from theoretical as well as from empirical side. 

\hspace{0.4cm} For deriving a risk measure with TTC, the heuristic assumption is made that the overall risk of a predicted collision 
at a time $s_E$ into the future decreases with increasing temporal distance to the event if nothing changes. 
When there is more time left until the incident, there is a larger chance for other things to happen (either voluntarily or 
unvoluntarily) which might lead to a different future evolution of the scene and to an avoidance of the event. 
An in-depth explanation is described later in Section \ref{SecRMSurv}. 
The most straightforward approach is then to calculate $s_E$ in order to get $R(t)$ according to 
\beq\label{EqRTTC}
R_{\mbox{\tiny TTC}}(t) \sim \left[\frac{1}{s_E}\right]^\alpha \ 
\eeq
with $\alpha>0$.
To avoid divergence and to fulfill the normalization conditions from Section \ref{SecRequirements} we introduce a small constant 
$\epsilon$ and a steepness constant $D_c$ so as to retrieve
\beq\label{EqRTTCPlusMod}
R_{\mbox{\tiny TTC}}(t) :=\left[\frac{\epsilon}{\epsilon+D_c \,s_E}\right]^{\alpha} \ .
\eeq \\
\vspace{-0.15cm}
\subsubsection{Extension for 2D operation} 
\hspace{10cm}

\hspace{0.4cm} Critical events like passing-by at high velocity can not be captured by Eq. (4). 
A logical refinement is given by considering not only collision events, but generally the future events
of highest criticality as a temporal reference. In case of collision risk, this leads to the time of closest proximity $s_E=\mbox{TTCE}$. We moreover separate the $\mbox{TTCE}$-dependent risk into two factors: One that handles the temporal decay in the usual 
way and another one that accounts for the increased spatial collision danger in case of high proximity. In this way, we obtain
\beq\label{EqRTTCPlusSpatial}
R_{\mbox{\tiny TTCE}}(t) :=\left(\frac{\epsilon}{\epsilon+D_c \,s_E}\right)^{\alpha}\, \exp\left\{-\frac{d_E^2}{2 \sigma^2}\right\} 
\eeq
with the Euclidean distance $d_E:=d(t+s_E)$ between the vehicles at the moment of the critical event and a Gaussian variance $\sigma^2$.

\hspace{0.4cm} In case of a collision, we would retain $d_E=0$ and the spatial term reduces to 1, so that we get back to 
$R_{\mbox{\tiny TTC}}(t)$. 
In case of a near-crash incident, the spatial term can be used to quantify how critical the incident was in terms of spatial 
proximity, accounting for uncertainty in the predicted positions.
As a further modification, we model the fact that spatial uncertainty in the predicted positions increases with larger prediction times 
$s_E$ (see also Section \ref{SecRMGauss}). This means that for events which lie further away in the future, we will consider larger $\sigma$'s 
with $\sigma(t+s_E):=D_c \,s_E$, so that 
\beq\label{EqGenTTCERisk}
R_{\mbox{\tiny TTCE}}(t) :=\underbrace{\left(\frac{\epsilon}{\epsilon+D_c \,s_E}\right)^{\alpha}}_{\mbox{\small temporal uncertainty}}\, 
\underbrace{\exp\left\{-\frac{d_E^2}{2D_{c}\,s_E}\right\}}_{\mbox{\small spatial uncertainty}} .
\eeq
With a prediction model for the trajectories $\xvec_{1,2}(t+s)$ of two TP's, we can directly calculate $s_E=\mbox{TTCE}$ as well as
$d_E$ at that time. 
In case of constant velocity assumptions
\beq
\xvec_{1,2}(t+s)=\xvec_{1,2}(t) + \vvec_{1,2}\,s
\eeq
we gain
\bea
&&d(t+s)=\lVert\Delta\xvec(t)+\Delta\vvec\,s\rVert \nonumber \\
&&\ \ =\sqrt{[\Delta\xvec(t)]^2+2\Delta\xvec\Delta\vvec\,s+(\Delta\vvec)^2\,s^2}
\eea
which has its minimum at
\beq\label{Eqse}
s_E=-\frac{\Delta\xvec(t)\,\Delta\vvec}{\lVert\Delta\vvec\rVert^2}
\eeq
and which reduces to TTC in the longitudinal case.
The distance of closest proximity is finally
\bea\label{EqDCP}
&&d_E = \sqrt{[\Delta \xvec(t)+\Delta\vvec \,s_E]^2}\nonumber \\
&&\ \ =\sqrt{\left\{\Delta \xvec(t)-\frac{\Delta\vvec\,[\Delta \xvec(t) \,\Delta \vvec]}{\lVert\Delta\vvec\rVert^2}\right\}^2}\nonumber \\
&&\ \ =\sqrt{\left[\frac{(\Delta\xvec(t) \times \Delta \vvec) \times \Delta\vvec}{\lVert\Delta\vvec\rVert^2}\right]^2} \nonumber \\
&&\ \ =\lVert\Delta\xvec(t)\rVert |\sin[\angle(\Delta\xvec(t),\Delta \vvec]| .
\eea

\begin{figure}[t!]
      \centering
      \vspace{0.32cm} 
      \resizebox{0.5\linewidth}{!}{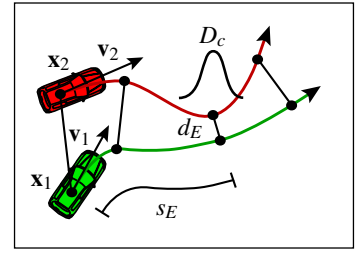}
      \captionsetup{justification=centering}
      \caption{\centering Collision risk prediction with TTCE.}      \label{fig:TTC2D}
\end{figure}

\hspace{0.4cm} As a result, using $d_E$ from Eq. (10) and inserting it into the spatial term of Eq. (6) we have gained a risk measure which generalizes TTC to the 
2D as well as to non-collision cases. Fig. \ref{fig:TTC2D} summarizes TTCE and its main variables.



\subsection{Gaussian Method} \label{SecRMGauss}

\hspace{0.4cm} As a second family of collision risk indicators, we consider approaches based on spatial occupancy probabilities\textsuperscript{ (1)}. For this purpose, 
the normalized probability densities for the respective spatial positions of two TP's indexed 1,2 are 
described by Gaussian functions
\beq
f_{1,2}(x)=\frac{1}{\sqrt{2\pi\sigma^2_{1,2}}} \, \exp\left[-\frac{(x-\mu_{1,2})^2}{2\sigma_{1,2}^2}\right]
\eeq
with the mean positions $\mu_{1,2}$ and variances $\sigma_{1,2}$.\footnote{For simplicity, we consider here the isotropic 1D case, 
but extensions to 2D and orientation are straightforward.}

\hspace{0.4cm} A collision at a position $x$ then occurs if {\em both} TP's coincide at the same position. Consequently, a way to quantify the likelihood of a 
coincidence/collision at a common position $x$ is
\beq
f_{c}(x):=f_1(x) f_2(x) \ \ .
\eeq

Because the product of two Gaussians is again a (non-normalized) Gaussian function, we get
\beq
f_{c}(x)=\frac{S_{c}}{\sqrt{2\pi\sigma^2_{c}}} \, \exp\left[-\frac{(x-\mu_{c})^2}{2\sigma^2_{c}}\right]
\eeq
with
\bea
&&\frac{1}{\sigma^2_{c}}=\frac{1}{\sigma^2_1}+\frac{1}{\sigma^2_2}, \hspace{0.3cm} \mu_{c}=\frac{\mu_1\sigma_1^2+\mu_2\sigma_2^2}{\sigma_1^2+\sigma_2^2} \ \ \mbox{and} \\
&&S_{c}=\frac{1}{\sqrt{2\pi(\sigma^2_{1}+\sigma^2_{2})}}\, \exp\left[-\frac{(\mu_1-\mu_2)^2}{2(\sigma^2_{1}+\sigma^2_{2})}\right] .
\eea

\hspace{0.4cm} The Gaussian position probability densities encompass the final positions of all possible trajectories that lead to points 
$x$ at a certain moment in time $t+s$.
The probability that the first TP, driving along its trajectory, is hit by the second TP is eventually given by spatially 
integrating $f_{c}(x)$ over all positions where the first TP can be
\beq\label{EqF}
P_E(s;t,\Delta t) \sim \int_{\infty} f_{c}(x) \, dx = S_{c} \ \ .
\eeq \\

\vspace{-0.25cm}

\subsubsection{From Collision Probability to Risk Measure}
\hspace{2cm}

\hspace{0.4cm} So how do we get from the collision probability factor (17) to a risk measure as a function of the time $t$ as introduced 
in Eq. (6)?   
First, we assume the moving TP's to follow a trajectory which undergoes certain variations in speed and 
heading. This accounts for mean positions through time $\mu_{1,2}(t+s)$ with spatial uncertainties $\sigma_{1,2}(t+s)$. For
$\sigma_{1,2}(t+s)$ a simple Brownian motion diffusion model with a linear increase of 
uncertainty starting at $\sigma_{1,2}(t)=0$ is used according to
\bea\label{EqDiffusionSigma}
\sigma^2_{1,2}(t+s):=D_{1,2}\, s. 
\eea
When we put Eq. (18) into (17), we obtain
\beq
P_E(s;t,\Delta t) \sim \frac{1}{\sqrt{2\pi D_{c}\,s}}\, \exp\left\{-\frac{[d(t+s)]^2}{2D_{c}\,s}\right\}
\eeq
with a joint diffusion constant $D_{c}:=D_1+D_2$ and an \afz{expected} distance $d(t+s):=\lVert\mu_1(t+s)-\mu_2(t+s)\rVert$.
Furthermore, to satisfy the requirements of Section \ref{SecRequirements} we add a small constant $\epsilon$ to the first term and thus gain 
\beq\label{EqPGauss}
P_E(s;t,\Delta t):=\left(\frac{\epsilon}{\epsilon+D_c\,s}\right)^{\alpha}\, \exp\left\{-\frac{[d(t+s)]^2}{2D_{c}\,s}\right\}
\eeq
with $\alpha=1/2$.

\hspace{0.4cm} Eq. (20) describes the probability that two TP's will be at the same position within a future interval around $[t+s]$, starting at $t$ and 
assuming Gaussian distributed positions around the TP's mean positions $\mu_{1,2}$. It can be seen that for larger prediction 
horizons $s$ the overall collision likelihood decreases, because of the larger uncertainty in the TP's positions.
The square-root dependency $\alpha=1/2$ is hereby a direct consequence of the Gaussian approach   
and the diffusion spread of the spatial probability densities. However, in a scene with allowed and non-allowed areas (road / non-road)
every time a TP leaves the road area it will already deviate from its \afz{normal} prediction, so that the collision probability will 
exhibit a faster decay with $s$.


\hspace{0.4cm} From Eq. (20) we can now distill a risk measure. As a conservative approach, we extract the maximal encountered future event 
probability as a risk indicator with
\beq
s_E:=\mbox{argmax}_s \, P_E(s;t,\Delta t)
\eeq
and retrieve for the Gaussian risk indicator
\beq
R_{\mbox{\tiny Gauss}}(t):=P_E(s_E;t,\Delta t)
\eeq
or rather
\beq\label{EqGenGaussRisk}
R_{\mbox{\tiny Gauss}}(t) :=\left(\frac{\epsilon}{\epsilon+D_c \,s_E}\right)^{\alpha}\, \exp\left\{-\frac{d_E^2}{2 \sigma^2}\right\} 
\eeq
whereas $d_E$ is again the Euclidean distance at the moment of the critical event $d_E=d(t+s_E)$.

\begin{figure}[tb]
      \centering
      \vspace{0.27cm}
      \resizebox{0.5\linewidth}{!}{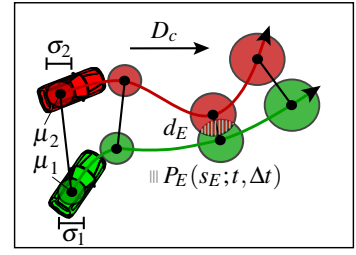}
      
      \caption{\centering Collision risk prediction with Gaussian method.} 
      \vspace{0.1cm}
      \label{fig:Gauss}
\end{figure}

\hspace{0.4cm} In Fig. \ref{fig:Gauss}, the increasing Gaussian position distributions are pictured, whereas the overlap between both at
the closest distance $d_E$ is taken for $R_{\mbox{\tiny Gauss}}(t)$.
Eq. (23) has the same form as Eq. (6). The difference is in the calculation of $s_E$, which in 
Section \ref{SecRMTTCE} occurs directly with TTCE, whereas here we estimate it via the maximized
$P_E(s;t,\Delta t)$. For small diffusion constants $D_c\rightarrow 0$ or imminent collisions $s_E \rightarrow 0$, the two 
approaches become equivalent. 

\subsection{Survival Analysis}\label{SecRMSurv}

\hspace{0.4cm} In Sections \ref{SecRMTTCE} and \ref{SecRMGauss}, we have seen two extensions of standard risk measures which try to circumvent some 
of their deficits. In particular, dealing with uncertainty over time and space as well as a generalization to 2D 
enables the use of risk measures in a broader range of situations. However, neither TTCE nor the Gaussian method provide a 
solid theoretical explanation, but rather motivated heuristics. 
In this Section, we describe a risk measure based on a grounded approach for the statistics of rare events and first passage time problems. 
This risk measure is also able to deal with uncertainties, but differently to the previously presented ones it provides full interpretability in terms of (normalized) probabilities and it considers additionally the situation history.


\hspace{0.4cm} Accident occurrences are modeled as a thresholding process based on a Poisson-like event probability \textsuperscript{ (12)}. 
For an exemplary vehicle, in a sufficiently small time interval of size $\Delta t$ the so-called {\bf instantaneous event probability} 
is characterized by an {\bf event rate} $\hat\tau^{-1}$ (units: events/sec) according to
\beq\label{EqInstEventProbClean}
\hat{I}_{\mbox{\tiny event}}(\Delta t):=\hat\tau^{-1}\Delta t \ \ .
\eeq
The term {\em instantaneous} event probability denotes the fact that this probability does not (yet) take the history into account.

\hspace{0.4cm} The {\bf survival probability} function $\hat{S}(t+s;t)$ indicates the probability that the vehicle will survive
from $t$ until $t+s$, i.e. that it will not be engaged in an event like an accident. 
From Eq. (24), we can directly derive the survival probability after a small time interval $\Delta t$ 
if the survival probability at $t'$ was $\hat{S}(t';t)$ 
\beq
\hat{S}(t'+\Delta t;t)=\hat{S}(t';t)\,[1-\hat\tau^{-1} \Delta t]
\eeq
so that with the starting condition $\hat{S}(t;t)=1$ we get 
\beq\label{EqSurv}
\hat{S}(t+s;t)=\mbox{exp}\{-\hat\tau^{-1}s\}
\eeq
which describes the {\bf homogeneous survival probability} for constant $\hat\tau^{-1}$.

\hspace{0.4cm} The real risk event modeling occurs by a proper parameterization and variation of the time-varying $\hat\tau^{-1}(t)$ resp.~its 
state-dependent analogous function $\tau^{-1}(\zvec_t)$ with $\hat\tau^{-1}(t)=\tau^{-1}(\zvec_t)$. In $\tau^{-1}(\zvec_t)$, 
we include all the risk factors with the context information in the state vector $\zvec_t$. Correspondingly, in dangerous situations 
the event rate will be higher than in harmless situations. 
For temporally varying $\hat\tau^{-1}(t)$, Eq. (26) modifies to
\beq\label{EqSurvivalProbClean}
\hat{S}(t+s;t)=\mbox{exp}\{-\int_o^s \hat\tau^{-1}(t+s') \, ds'\}\ \ .
\eeq
The states have been left out here for simplicity of the derivations. If we include them back, we acquire the state-dependent survival 
probability function
\beq\label{EqSurvivalProb}
S(t+s;t,\zvec_{t:t+s})=\mbox{exp}\{-\int_o^s \tau^{-1}(\zvec_{t+s'}) \, ds'\}
\eeq
which defines the probability that a vehicle survives during $[t,t+s]$ without being involved in a critial event and which depends 
on the entire state vector sequence $\zvec_{t:t+s}$. 

\vspace{0.1cm}

\subsubsection{From Survival Probability to Risk Measure}
\vspace{0.1cm}
\hspace{2cm}

\hspace{0.4cm} To quantify the risk of an accident between the ego-car and another car, the (time-varying!) {\bf spatial risk instantaneous collision 
event rate} $\tau_c^{-1}(\zvec)$ is modelled by
\beq\label{EqEventRateMulTP}
\tau_{\mbox{\tiny coll}}^{-1}(\zvec)=\tau_{{\mbox{\tiny coll}},0}^{-1}\,e^{-\beta_{\mbox \tiny coll}\,|\xvec_1-\xvec_2|}
\eeq
with constants $\tau_{\mbox{\tiny coll,0}}^{-1}$ and $\beta_{\mbox{\tiny coll}}$. While the scale factor $\tau_{\mbox{\tiny coll,0}}^{-1}$
is chosen so that $\hat{I}_{\mbox{\tiny event}}(\Delta t, \zvec)$ 
of each other car approaches 1 at collision, the steepness factor $\beta_{\mbox{\tiny coll}}$ is used to model 
the position uncertainty originated from several possible sources like sensor inaccuracy, state prediction errors, unexpected driver 
behavior or unknown vehicle sizes. With $\beta_{\mbox{\tiny coll}}$, closer proximity leads to higher $\tau_{\mbox{\tiny coll}}^{-1}(\zvec)$
and accordingly to a reduced probability of \afz{surviving}.

\hspace{0.4cm} The approach is consistently extensible to different types and sources of risk by using a composed event rate like
\bea\label{EqEventRateMulRisks}
\tau^{-1}(\zvec)&:=&\tau^{-1}_0+\tau^{-1}_{\mbox{\tiny coll}}(\zvec)+\tau^{-1}_{\mbox{\tiny ctrl}}(\zvec)+... \nonumber \\
&=& \tau^{-1}_0 + \tau^{-1}_{\mbox{\tiny crit}}(\zvec)
\eea
which can comprise the terms related to critical events, such as vehicle-to-vehicle collisions $\tau^{-1}_{\mbox{\tiny coll}}$, losing control 
in curves $\tau^{-1}_{\mbox{\tiny ctrl}}$ and others. 
The {\bf escape rate} $\tau^{-1}_0$ plays a special role: It contains all (unknown and non-critical) \afz{escape} events which might 
lead to the case that the currently predicted future gets invalid and later critical events lying further away in the future 
cannot occur any more. For instance, if we assume constant velocity in the prediction and the collision will occur at $\mbox{TTC}=\unit[10] {s}$, each disturbance, 
voluntary or involuntary action away from the constant velocity assumptions that occurs within $[t,t+\mbox{TTC}]$ 
will prevent the collision to happen (the driver will \afz{escape} from the collision).

\hspace{0.4cm} By multiplying the instantaneous event probability from Eq. (24) with the survival probability from Eq. (28),
the {\bf event probability} can be calculated as 
\beq\label{EqEventProb}
E(s;t,\Delta t, \zvec_{t:t+s}) =I_{\mbox{\tiny event}}(\Delta t,\zvec_{t+s})\,S(t+s;t,\zvec_{t:t+s}) .
\eeq
This is the probability that any type of event (escape event or critical) will happen in an interval of length $\Delta t$ around $t+s$, 
given a state vector history $\zvec_{t:t+s}$ if we start observations at $t$ and no event happened during $[t,t+s]$.
\begin{figure}[tb]
      \centering
      \vspace{0.62cm} 
      \resizebox{0.5\linewidth}{!}{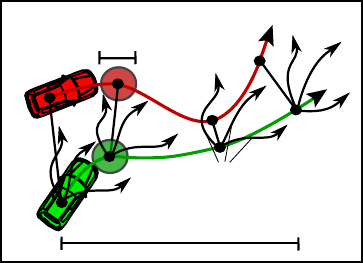}
      
      \caption{\centering Collision risk prediction with survival analysis.} 
      \vspace{0.1cm}
      \label{fig:SA}
\end{figure}
Correspondingly, the {\bf event density} (i.e. the probability of events per time unit) after a time $s$ starting at $t$ is given by
\bea\label{EqEventDens}
&&e(t+s;t,\zvec_{t:t+s}):=E(s;t,\Delta t,\zvec_{t:t+s})/\Delta t \nonumber \\
&&\ \ \ \ =\tau^{-1}(\zvec_{t+s})\,S(t+s;t,\zvec_{t:t+s})
\eea
and the total {\bf time-accumulated event probability} (i.e. the probability that any of the events will happen during $[t,t+s]$) by
\beq
A(s;t,\zvec_{t:t+s}):=\int_o^s e(s';t,\zvec_{t:t+s'})\,ds' \ \ .
\eeq

\hspace{0.4cm} As a risk measure, we want to take the time-accumulated probability of the critical events only. If we separate the critical 
events from the escape events (which \afz{avoid} the accidents), we find out that the event density
\bea
&&e(s;t,\zvec_{t:t+s})=\tau^{-1}(\zvec_{t+s})\,S(t+s;t,\zvec_{t:t+s}) \nonumber \\
&&\ \ =[\tau^{-1}_0 + \tau^{-1}_{\mbox{\tiny {crit}}}(\zvec_{t+s})]\,S(t+s;t,\zvec_{t:t+s}) \nonumber \\
&&\ \ := e_{0}(s;t,\zvec_{t:t+s}) + e_{\mbox{\tiny crit}}(s;t,\zvec_{t:t+s})
\eea
also separates into two respective terms. Consequently, for the time-accumulated event probability
\beq
A(s;t,\zvec_{t:t+s})=A_0(s;t,\zvec_{t:t+s})+A_{\mbox{\tiny {crit}}}(s;t,\zvec_{t:t+s}) \ \ .
\eeq
holds true.
The first term quantifies the overall future probability that critical events will be avoided during $[t,t+s]$, whereas the second 
term expresses the future probability of getting involved in a critical event during the same time span.

\hspace{0.4cm} Since we are interested in the estimation of the time-accumulated future risk, we use 
\beq 
P_E(s;t,\Delta t):=e_{\mbox{\tiny {crit}}}(s;t,\zvec_{t:t+s})\Delta t
\eeq
as our time-resolved differential risk measure. The integral risk measure then is
\beq
R(t)=A_{\mbox{\tiny {crit}}}(\infty;t,\zvec_{t:t+\infty})
\eeq
It can be shown that the time-accumulated event probability integrates to 1, which means
\beq
A(\infty;t,\zvec_{t:t+\infty})=1\ \ .
\eeq
This is a sensible and necessary normalization condition, because for an infinite future a critical event will happen with certainty 1.

\hspace{0.4cm} Using this information, we arrive at
\bea
&&R_{\mbox{\tiny SA}}(t)=1-A_0(\infty;t,\zvec_{t:t+\infty}) \\
&&\ \ =1-\underbrace{\tau^{-1}_0\int_o^{\infty} S(t+s;t,\zvec_{t:t+s'}) \,ds'}_{\mbox{\tiny Overall escape probability}}
\vspace{-2cm}
\eea

for our final risk measure. It contains the overall probability of escaping a future critical event and its complement 
quantifies the overall probability of engaging in a future critical event. Besides, it is well-behaved in the limiting cases. The risk measure 
automatically approaches $0$ if there are no critical events present, since then the overall escape probability approaches 1 and for an 
imminent critical event, the escape probability reaches 0 as there is no time left for any type of escape events in terms of avoidance 
behavior or similar. Fig. \ref{fig:SA} illustrates $\tau^{-1}_0$ as trajectory alternatives changing $\zvec_{t:t+s}$, the
positional uncertainty $\beta_{\mbox{\tiny coll}}$ and the procedure of integrating the survival function $S(t+s;t)$ over the predicted time $s$
which among others depends on $\tau_{\mbox{\tiny coll}}^{-1}$.

\vspace{0.2cm}
\section{Simulation Results} \label{SecSimRes}

\subsection{Evaluation of Single Longitudinal and Intersection Scenario} \label{SecEvalSingle} 

\hspace{0.4cm} For a quantitative comparison, we applied the presented three risk measures to real crash cases taken from the German
In-Depth Accident Study (GIDAS) dataset\textsuperscript{ (14)}. The GIDAS Pre-Crash-Matrix contains reconstructed trajectories of two 
TP's involved in an upcoming 
collision for longitudinal and intersection scenarios on average $t=$ -$\unit[5.5]{s}$ ahead. In order to also quantify the risk 
for near-crash cases, we additionally changed the course of the scene evolution. In the longitudinal car-following example, 
the path of one TP was shifted laterally in such a way that the minimal distance is $d_E= \unit[7]{m}$ instead 
of $d_E=\unit[0]{m}$. 
There is no collision anymore, but a close passing. In the intersection example, we used for one TP 
the Foresighted Driver Model (FDM)\textsuperscript{ (15)}, which changes the velocity profile $v(t)$ of its trajectory. Accordingly, the TP 
decelerates and lets the other TP pass to avoid an accident.

\hspace{0.4cm} At every timestep $t$ in the simulation, a constant velocity model is used for both TP's to predict the distance $d_{t:t+s}$
for future times $s$. The prediction horizon is set as $s_H>\unit[5.5]{s}$ so that the time of crash or near-crash 
$s_E$ is within the prediction interval. The sequence $d_{t:t+s}$ represents the input for all three risk measures. In this way,
they have the same prerequisites and if the real trajectory violates the constant velocity assumption with 
an acceleration $a(t)\neq \unit[0]{m/s^2}$, all risk measures are equally impacted. Furthermore, we selected the parameter
$D_c$ of TTCE and the Gaussian method and $\tau^{-1}_0$ as well as $\beta_{\mbox{\tiny coll}}$ for the survival analysis 
such that for the maximal risk value of each $R_{max}>0.5$ holds true in the near-crash case.

\begin{figure}[tb]
      \centering
      \parbox{1.03\linewidth}{
	\resizebox{\linewidth}{!}{
	  \centering
	  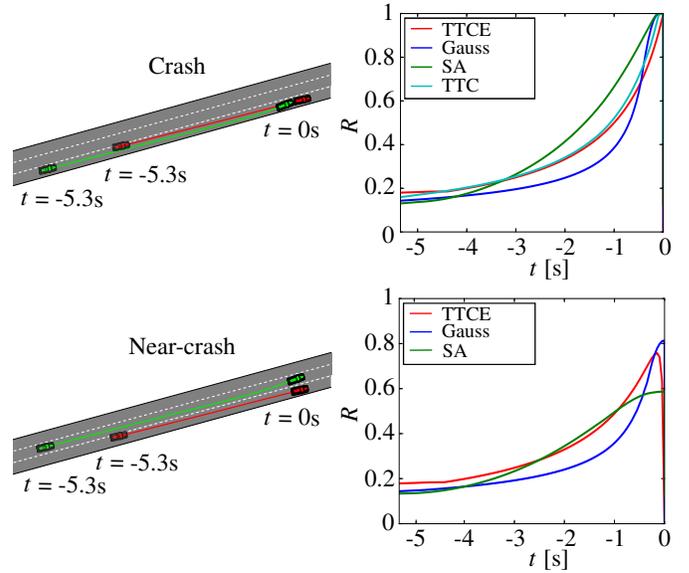
	 }
      }
      \caption{\centering Comparison of risk measures for longitudinal scenario with simulation view and their respective curves. Top: Crash case. Bottom: Near-crash case.} 
      \vspace{0.1cm}
      \label{fig:lon}
\end{figure}

\begin{figure}[tb]
      \vspace*{-0.5cm}
      \centering
      \parbox{1.03\linewidth}{
	\resizebox{\linewidth}{!}{
	  \centering
	  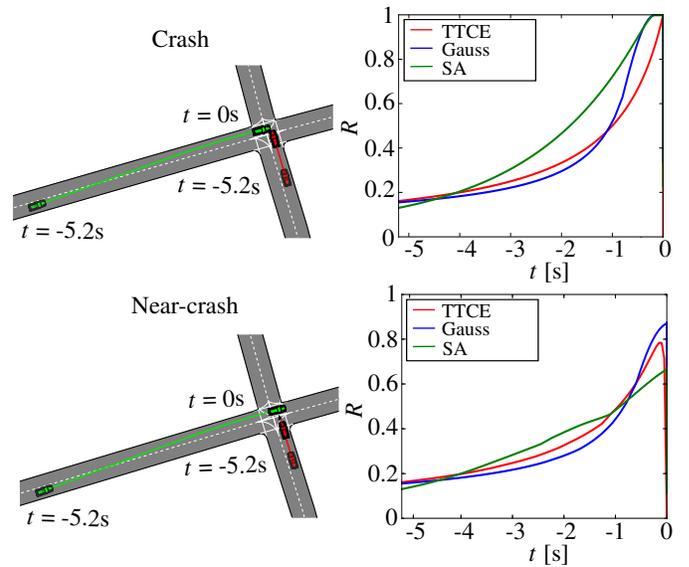
	}
      }
      \caption{\centering Comparison of risk measures for intersection scenario.} 
      \label{fig:inter}
\end{figure}

\hspace{0.4cm} Fig. \ref{fig:lon} shows the simulation view of a longitudinal crash and near-crash case starting from time 
$t=$ -$\unit[5.3]{s}$ until the point of maximal criticalty $t_E=\unit[0]{s}$ and plots of the risk measures $R_{\mbox{\tiny TTCE}}(t)$, 
$R_{\mbox{\tiny Gauss}}(t)$ and $R_{\mbox{\tiny SA}}(t)$. Equally, Fig. \ref{fig:inter} summarizes the simulation
results for an intersection scenario with the interval $t=[$-$\unit[5.2]{s},\unit[0]{s}]$. In both scenarios $R_{\mbox{\tiny SA}}(t)$ approaches 1 faster in the crash case
and has a lower $R_{max}$ in the near-crash case. Compared to $R_{\mbox{\tiny TTCE}}(t)$, the performance 
of $R_{\mbox{\tiny Gauss}}(t)$ is higher, but its robustness lower. The reason lies in the square-root dependency 
$\alpha=1/2$ in $R_{\mbox{\tiny Gauss}}(t)$ as opposed to $\alpha=1$, which makes its curve shape flatter at $t<t_E$ and steeper close to 
$t_E$. Lastly, only for the longitudinal crash case $R_{\mbox{\tiny TTC}}(t)$ can be calculated with the help of Eq. (4). 
It resembles $R_{\mbox{\tiny TTCE}}(t)$, but has slightly higher values because of the missing
spatial uncertainty term from Eq. (6).

\subsection{Statistical Analysis for Multiple Scenarios} 

\begin{figure}[tb]
      \centering
      \parbox{1.03\linewidth}{
	\resizebox{\linewidth}{!}{
	  \centering
	  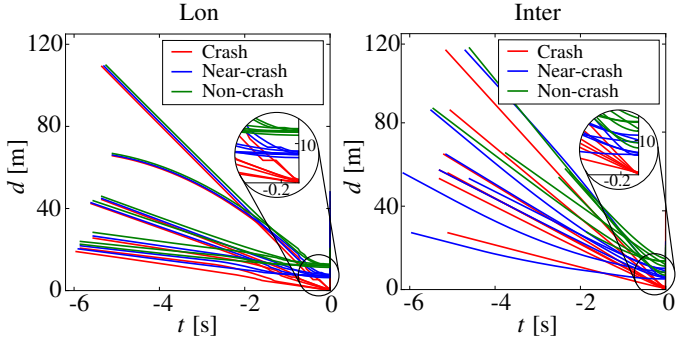
	}
      }
      \caption{\centering Variance in distance sequences of extended GIDAS dataset. Left: Longitudinal scenario. Right: Intersection scenario.} 
      \label{fig:dataset}
      \vspace{0.3cm}
\end{figure}

\hspace{0.4cm} After demonstrating the general behavior of the risk measures, we now test them on a set of
42 scenarios. The set consists of 7 longitudinal and 7 intersection scenarios from GIDAS, which have not only
a crash and near-crash case, but also a non-crash case. In the longitudinal example, we moved the path
of one TP laterally so that $d_E= \unit[12]{m}$ is reached and in the intersection samples, we set constant velocities
for both TP with the result that they pass the intersection $|\Delta t| =2s$ away from each other.   
In Fig. \ref{fig:dataset} the distance sequences $d(t)$ are pictured for all 42 test instances, which ranges from
$\unit[120]{m}$ to $\unit[0]{m}$. The pictured graphs $d(t)$ should not be confused with the predicted distances
$d_{t:t+s}$, which are calculated at every time step $t$ with the constant velocity model and which act as the input
of the risk measures.

\hspace{0.4cm} When a risk measure crosses a threshold of $R_{th}=0.7$, we define it as having detected a crash.
An optimal risk measure has an early detection time $t_d$ of the crash in the crash cases and has no false-positive 
detections $FP$ in the near- and non-crash cases, which occurs if $R_{max}>R_{th}=0.7$ takes effect. 
In Table \ref{tab:Statistics}, the averaged values of $t_d$ and $R_{max}$ and the accumulated $FP/N$ of each 
risk measure are listed for the 42 test samples. Furthermore, the variances $\sigma_\text{\fontsize{7}{7}\selectfont  $t$}$ and $\sigma_R$   
show the spread of the results. As anticipated from the last Section \ref{SecEvalSingle}, the survival analysis has the highest $|t_d|$ and the lowest
$R_{max}$ and $FP/N$. The gaussian method exhibits a larger $|t_d|$ than TTCE, but in the near-crash case 
$R_{max}$ and $FP/N$ is smaller in TTCE. 
In general, a crash is detected later and there are more false detections
in intersection compared to longitudinal scenarios. That is because a real trajectory with $a(t)\neq \unit[0]{m/s^2}$ results
into a more parabolic curve of $d(t)$ in intersection scenarios (see Fig. \ref{fig:dataset}), which in turn causes 
higher errors in $d_{t:t+s}$ due to the constant velocity assumption.
Finally, TTC is included as a standard method in the ADAS and AD community to measure collision risk. However,
in the evaluated cases the performance of our developed three risk measures has proven to be better and they work 
in a broader range of scenarios, i.e. in intersection scenarios as well as in near- and non-crash cases.
\footnote{Remark: In Table \ref{tab:Statistics}, TTC has a higher (meaning better) $|t_d|$ compared to TTCE. This might
seem unexpected, because we derived TTCE to be a generalization of TTC. However, the TTCE parametrization was optimized to work in all cases (crash, near-crash and non-crash), whereas the TTC implementation specializes on longitudinal crash cases only.}

\vspace{0.2cm}
\section{Discussion and Outlook} \label{SecDiscOut}

\hspace{0.4cm} In this work, we first extended the heuristic risk measures TTC and Gaussian method to be able to deal with 
temporal and spatial uncertainty and to fulfill the normalization requirements for a general framework of collision risk 
prediction. For TTC, we additionally derived a 2D version called TTCE. In a second step, we introduced 
a theoretically justified survival analysis which separates events into critical and escape events and 
calculates a risk measure by integrating the probability to survive over the predicted time. 
Each of the three methods precede a prediction step of the scene evolution with the output of predicted distances.

\hspace{0.4cm} In simulations of longitudinal and intersection scenarios, the survival analysis resulted in having the highest performance
in terms of early detection time in the crash cases as well as robustness with less false-positive detections in the near- and non-crash cases.
The Gaussian method and TTCE have similar accuracy, which is reasonable since their underlying equations could be shown to be 
also very similar. Furthermore, both can be seen as an approximation of the survival analysis without history consideration. 

\hspace{0.4cm} Procedures to validate ADAS and AD have to be developed beyond existing safety tests driving
millions of miles\textsuperscript{ (16)}. 
Since TTC only works for longitudinal scenarios and has moderate performance, the survival analysis 
is more suitable. 
In future work, we want to develop a risk indicator quantifying the entire experienced risk for  
long test drives and arbitrary driving situations.

\hspace{0.4cm} Many motion planning methods act on TTC-based risk measures. Enhancements have been made in the FDM\textsuperscript{ (14)},
which uses gradient descent on a modified TTCE with predicted trajectories from a constant velocity model.     
As an alternative, the survival analysis could be employed on multiple trajectories of arbitrary velocity profiles. 
The trajectory with the lowest risk is eventually chosen as the planned behavior. Such an approach that relies on sampling shows to be 
promising and needs to be examined.

\begin{table}[t!]
\caption{\hspace*{-3.9cm}\mbox{\centering Statistics of risk measures for extended GIDAS dataset.}} 
\begin{center}
\begin{tabular}{|c|p{0.8cm}|p{0.65cm}|p{0.65cm}|p{0.5cm}|p{0.6cm}|p{0.5cm}|p{0.6cm}|} 
\cline{3-8}
\multicolumn{2}{c|}{} & \multicolumn{2}{c|}{Crash} & \multicolumn{2}{c|}{Near-crash} & \multicolumn{2}{c|}{Non-crash} \\
\cline{3-8}
\multicolumn{2}{c|}{} & Lon & Inter & Lon & Inter & Lon & Inter \\
\hline
\multirow{3}{*}{TTCE}  & $t_d$ [s] & -$0.47$ & -$0.45$ & $-$ & $-$ & $-$ & $-$ \\
\cline{2-8}
& $\sigma_\text{\fontsize{7}{7}\selectfont  $t$}$ [s] & $0.05$ & $0.01$ & $-$ & $-$ & $-$ & $-$ \\ 
\cline{2-8}
 & $R_{max}$ & $-$ & $-$ & 0.78 & 0.75 & 0.62 & 0.63 \\
 \cline{2-8}
& $\sigma_R$ & $-$ & $-$ & 0.02 & 0.08 & 0.01 & 0.08 \\
\cline{2-8}
 & $FP/N$ & $-$ & $-$ & 7/7 & 4/7 & 0/7 & 2/7 \\
\hline
\multirow{3}{*}{Gauss}  & $t_d$ [s] & -1.36 & -0.85 & $-$ & $-$ & $-$ & $-$ \\
\cline{2-8}
& $\sigma_\text{\fontsize{7}{7}\selectfont  $t$}$ [s] & $0.75$ & $0.39$ & $-$ & $-$ & $-$ & $-$ \\
\cline{2-8}
 & $R_{max}$ & $-$ & $-$ & 0.84 & 0.82 & 0.53 & 0.57 \\
\cline{2-8}
& $\sigma_R$ & $-$ & $-$ & 0.03 & 0.11 & 0.03 & 0.16 \\
\cline{2-8}
 & $FP/N$ & $-$ & $-$ & 7/7 & 6/7 & 0/7 & 2/7 \\
\hline
\multirow{3}{*}{SA}  & $t_d$ [s] & -1.46 & -1.14 & $-$ & $-$ & $-$ & $-$ \\
\cline{2-8}
& $\sigma_\text{\fontsize{7}{7}\selectfont  $t$}$ [s] & $0.48$ & $0.23$ & $-$ & $-$ & $-$ & $-$ \\
\cline{2-8}
 & $R_{max}$ & $-$ & $-$ & 0.63 & 0.63 & 0.35 & 0.42 \\
\cline{2-8}
& $\sigma_R$ & $-$ & $-$ & 0.04 & 0.12 & 0.02 & 0.15 \\
\cline{2-8}
 & $FP/N$ & $-$ & $-$ & 0/7 & 3/7 & 0/7 & 0/7 \\
\hline
TTC & $t_d$ [s] & -0.76 & $-$ & $-$ & $-$ & $-$ & $-$ \\
\cline{2-8}
& $\sigma_\text{\fontsize{7}{7}\selectfont  $t$}$ [s] & $0.16$ & $-$ & $-$ & $-$ & $-$ & $-$ \\
\hline

\end{tabular}
\label{tab:Statistics}
\end{center}
\vspace{0.2cm}
\end{table}

\hspace{0.4cm} At last, it is possible to improve the collision risk estimation in general by separating the distance calculation into longitudinal and lateral components along the ego path and weighting the components differently. 
From our simulation experiences, we expect that this will render the method more sensitive yet further reducing the number of false positives in near-crash cases.

\section*{Acknowledgments}
This work has been supported by the European Union’s Horizon 2020 project \textit{VI-DAS}, under the grant agreement number 690772.

\section*{References}

\hskip-0.22cm
\begin{tabular}{p{0.26cm} p{7.74cm}} 
(1)&US Department of Homeland Security: Risk Management Fundamentals, pp. 19-21 (2011). \\
 (2) & S. Lefevre, D. Vasquez, and C. Laugier: A Survey on Motion Prediction and Risk Assessment for Intelligent Vehicles, Robomech Journal, Volume 1, Issue 1, pp. 1-14 (2014). \\
 (3) & M. Bojarski, D. Del Testa, D. Dworakowski, B. Firner, 
and et al.: End to End Learning for Self-Driving
Cars, Computing Research Repository, abs/1604.07316 (2016).  \\
 (4) & S. Klingelschmitt, M. Platho, H.-M. Groß, V. Willert, and J. Eggert: Combining Behavior and Situation Information for Reliably Estimating Multiple Intentions, Intelligent Vehicles Symposium, pp. 388-393 (2014).  \\
 (5) & R. Van der Horst: Time-To-Collision as a Cue for Decision-Making in Braking, Vision in Vehicles III, pp. 19-26 (1991).  \\
 (6) & H. Winner: Grundlagen, Komponenten, und Systeme f\"ur aktive Sicherheit und Komfort, Handbuch Fahrerassistenzsysteme, pp. 522–542 (2014). \\
 (7) & J. Ward, G. Agamennoni, S. Worrall, and E. Nebot: Vehicle Collision Probability Calculation for General
Traffic Scenarios Under Uncertainty, Intelligent Vehicles Symposium, pp. 986-992 (2014).  \\

\end{tabular}

\hskip-0.22cm
\begin{tabular}{p{0.26cm} p{7.74cm}} 
 (8) & F. Damerow and J. Eggert: Predictive Risk Maps, Intelligent Transportation Systems Conference, pp. 703-710 (2014).  \\
 (9) & S. Patil, J. Van den Berg, and R. Alterovitz: Estimating Probability of Collision for Safe Motion Planning
under Gaussian Motion and Sensing Uncertainty, International Conference on Robotics and Automation, pp. 3238 - 3244 (2012).  \\
 (10) & A. Lambert, D. Gruyer, and G. Saint Pierre: A fast Monte Carlo Algorithm for Collision Probability Estimation, Control, Automation, Robotics and Vision Conference, pp. 406-411 (2008). \\
 (11) & W. Wachenfeld, P. Junietz, R. Wenzel, and H. Winner: The Worst-Time-To-Collision Metric for Situation
Identification, Intelligent Vehicles Symposium, pp. 729-734 (2016). \\
 (12) & J. Eggert: Predictive Risk Estimation for Intelligent ADAS Functions, Intelligent Transportation Systems Conference, pp. 711-718 (2014). \\
\end{tabular}

\newpage
\hskip-0.22cm
\begin{tabular}{p{0.26cm} p{7.74cm}} 
 (13) & J. Eggert and T. Puphal: Continuous Risk Measures for ADAS and AD, International Symposium on Future Active Safety Technology, (2017). \\
  (14) & C. Erbsmehl: Simulation of real crashes as a method for estimating the potential benefits of advanced safety
technologies. In Technical Conference on the Enhanced Safety of Vehicles, No. 09-0162, (2009). \\ 
   (15) & J. Eggert, F. Damerow, and S. Klingelschmitt: The Foresighted Driver Model, Intelligent Vehicles Symposium, pp. 322-329 (2015).
 \\ 
    (16) &  D. Zhao and H. Peng: From the Lab to the Street: Solving the Challenge of Accelerating Automated Vehicle Testing, Computing Research Repository, abs/1707.04792 (2017). 
 \\ 
\end{tabular}

\end{document}